\documentclass[conference,letterpaper]{IEEEtran}

\usepackage{graphicx}%graph package
\usepackage{amsmath}%matchematic package
\usepackage{amssymb}
\usepackage{stfloats}%adjust the table position

\usepackage{xcolor}
\usepackage{booktabs}
\usepackage{tabularx}
\usepackage{url}
\usepackage{algorithm}  
\usepackage{algpseudocode}  

\IEEEoverridecommandlockouts
\usepackage{multirow}
\usepackage[left=0.71in,top=0.94in,right=0.71in,bottom=1.18in]{geometry}
\begin{document}
\title{\LARGE \bf Human-Aware Robot Navigation via Reinforcement Learning with Hindsight Experience Replay and Curriculum Learning}

\author{Keyu Li, Ye Lu, and Max Q.-H. Meng$^{*}$, \textit{Fellow}, \textit{IEEE}
\thanks{This work was partially supported by National Key R\&D program of China with Grant No. 2019YFB1312400, Shenzhen Key Laboratory of Robotics Perception and Intelligence (ZDSYS20200810171800001), Hong Kong RGC CRF grant C4063-18G, Hong Kong RGC GRF grant \#14211420 and Hong Kong RGC GRF grant \#14200618 awarded to Max Q.-H. Meng.}
\thanks{K. Li and Y. Lu are with the Department of Electronic Engineering, The Chinese University of Hong Kong, Hong Kong, China (e-mail: kyli@link.cuhk.edu.hk; luyyy@link.cuhk.edu.hk).}%
\thanks{Max Q.-H. Meng is with the Department of Electronic and Electrical Engineering of the Southern University of Science and Technology in Shenzhen, China, on leave from the Department of Electronic Engineering, The Chinese University of Hong Kong, Hong Kong, and also with the Shenzhen Research Institute of the Chinese University of Hong Kong in Shenzhen, China (e-mail: max.meng@ieee.org).}%
\thanks{$^{*}$Corresponding author.}%
}

\maketitle
\begin{abstract}
	In recent years, the growing demand for more intelligent service robots is pushing the development of mobile robot navigation algorithms to allow safe and efficient operation in a dense crowd. Reinforcement learning (RL) approaches have shown superior ability in solving sequential decision making problems, and recent work has explored its potential to learn navigation polices in a socially compliant manner. However, the expert demonstration data used in existing methods is usually expensive and difficult to obtain. In this work, we consider the task of training an RL agent without employing the demonstration data, to achieve efficient and collision-free navigation in a crowded environment. To address the sparse reward navigation problem, we propose to incorporate the hindsight experience replay (HER) and curriculum learning (CL) techniques with RL to efficiently learn the optimal navigation policy in the dense crowd. The effectiveness of our method is validated in a simulated crowd-robot coexisting environment. The results demonstrate that our method can effectively learn human-aware navigation without requiring additional demonstration data.
\\
\end{abstract}

% key words

\section{INTRODUCTION}

With the dramatic development of machine intelligence in recent years, robots have been expected to work in social space shared with humans in a large number of real-life tasks \cite{ferdowsi2018robust}\cite{morales2018passenger}. Navigation in a dense crowd efficiently and safely is an important yet challenging task for the operation of the mobile robots, which requires the robot to have the ability to understand human behavior and take actions under cooperative rules.

\begin{figure}[tb]
      \centering
      \includegraphics[scale=1.0,angle=0,width=0.49\textwidth]{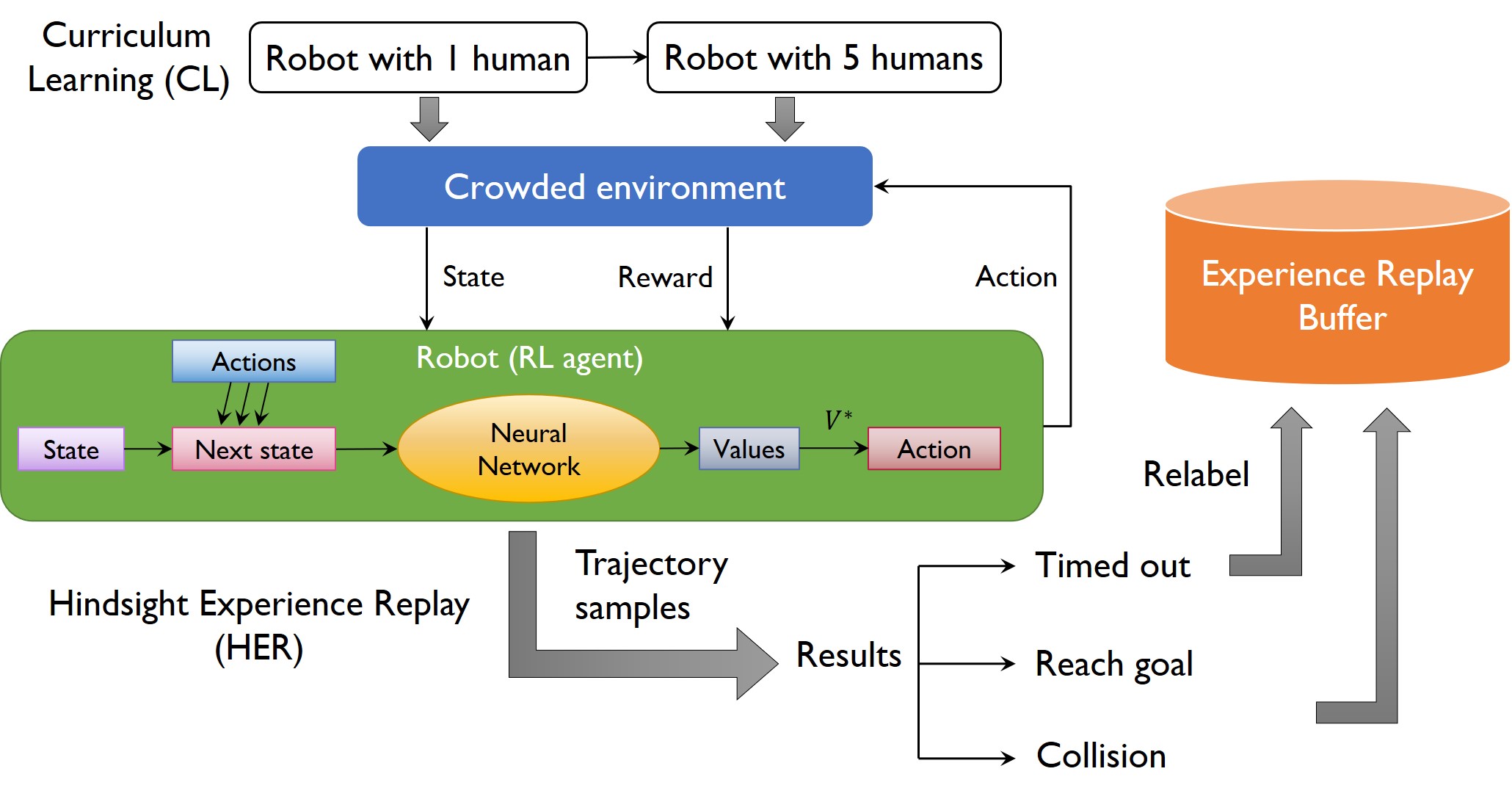}
      \caption{Overview of our proposed framework that uses a deep reinforcement learning (RL) agent trained with hindsight experience replay (HER) techniques and curriculum learning (CL) to achieve human-aware robot navigation in crowded environments.}
      \label{overview}
\end{figure}

\subsection{Related Work}

\subsubsection{Hand-crafted interaction models} 
Earlier attempts have been made to develop some interaction models to help the robot understand social behaviors and avoid obstacles in navigation. One representative work is the social force model \cite{helbing1995social}, which analyzes crowd behavior based on the modelling of interaction forces. This model has been extensively used for robot navigation in both simulation and real-world applications \cite{ferrer2013social}. Interacting Gaussian processes model \cite{trautman2010unfreezing} is another method to estimate and predict crowd interaction pattern from recorded data by describing a probabilistic interaction between multiple navigating entities. However, these models usually rely on some hand-crafted functions and parameters, which are limited to specific scenarios.

\subsubsection{Imitation learning approaches} 
Imitation learning (IL) has also been utilized to make the agents learn optimal navigation policies from expert demonstrations. By directly using the provided demonstrations, navigation polices can be trained to directly output the robot's movement to imitate the human behavior \cite{tai2018socially}\cite{kollmitz2020learning}. However, these methods are highly dependent on the quantity and quality of expert demonstrations to achieve a good performance. 

\subsubsection{Reinforcement learning approaches} 
In recent years, reinforcement learning (RL) has been extensively studied in various applications such as medical imaging \cite{li2021autonomous}. In crowd-aware robot navigation, there have been some works \cite{chen2017socially}\cite{cadrl}\cite{everett2018motion} that use deep RL to learn optimal navigation policies in a socially cooperative manner, by implicitly encoding the interactions and cooperation among agents. However, these methods modelled the crowd impact by aggregating the pairwise interactions and focusing on one-way interactions from humans to the robot, which would limit the performance of cooperative planning in complex and crowded scenes. To overcome these limitations, a recent study \cite{sarl} proposed to rethink human-robot pairwise interactions and integrated a self-attention module in the deep RL framework, to achieve better navigation performance. Li et al. \cite{sarlstar} extended this method to achieve crowd-aware navigation in indoor environments and demonstrated the feasibility of their method in real-world experiments.
%However, it trains the optimal navigation policy assuming that the environment model is known. It is a model-based method, which is not the case usually in the real world. The results can be ideal to some extent. In our work, based on this RL framework, we will study the performance of different popular algorithms  and propose to apply model-free methods to achieve better and more realistic results.

However, all the aforementioned methods initialize the policy of the RL agent with IL by using demonstration data collected with an expert policy, in order to accelerate the learning process \cite{chen2017socially}--\cite{sarlstar}. 
In spite of the benefits of using expert demonstrations for pre-training of the agent, there are some limitations in practice. First, these methods initialize the policy of the RL agent as the human policy in the simulation environment, which would greatly simplify the problem and may not be applicable in reality. Second,  a large dataset of high-quality human demonstrations in the real-world navigation is difficult and expensive to obtain. Therefore, in this work, we investigate the human-aware navigation problem by using RL approaches under the assumption that no expert demonstration data is provided, in order to explore more versatile algorithms to achieve both efficient and collision-free navigation in a crowded scene. 

\subsection{Overview}
An overview of our method can be seen in Fig. \ref{overview}. Adopting the same socially attentive deep RL framework as described in \cite{sarl}\cite{sarlstar}, we forgo the use of the expert demonstration data for initialization of the agent, and propose to integrate the hindsight experience replay (HER) technique to address the sparse reward navigation problem. Moreover, curriculum learning (CL) is used to transfer the knowledge gained in a simpler environment (containing the robot and $1$ human) to a more complex environment (containing the robot and $5$ humans).

The remainder of this paper is structured as follows. Section~II gives the formulation of the crowd navigation problem. Section~III introduces the details of the proposed methods, before experiment results are discussed in Section~IV. Finally, we draw some conclusions in Section~V.

\section{PROBLEM FORMULATION}
Suppose there is a human-robot coexisting environment where a robot and $n$ humans are navigating in a 2D workspace denoted as the X-Y plane. The humans and the robot have their own destinations, and the goal of the robot is to navigate to its destination without colliding with humans.  We can formulate this problem as a sequential Markov Decision Process (MDP) model in the framework of reinforcement learning. It is assumed that humans do not avoid the robot during the navigation, and each agent can be simplified as a moving circle. We assume that the position, velocity and radius of each agent can be observed by other agents, which are denoted as $\boldsymbol{p} = [p_{x}, p_{y}]$, $\boldsymbol{v} = [v_{x}, v_{y}]$ and $r$, respectively. The goal position $\boldsymbol{g} = [g_{x}, g_{y}]$ and preferred speed $v_{pref}$ cannot be observed by other agents. A robot-centric frame defined in \cite{cadrl} is adopted here to make the state representation more general and versatile, where the origin of the frame is set at the current position of the robot at time $t$ $\boldsymbol{p}_t$, and X-axis points at its goal position $\boldsymbol{g}$. Let $d_{g}$ denote the distance from $\boldsymbol{p}_t$ to $\boldsymbol{g}$, and $d^{i}$ denote the distance from $\boldsymbol{p}_t$ to the position of the $i$-th human at time $t$ $\boldsymbol{p}_t^{i}$. After transformation, the state of the robot at time $t$ $\boldsymbol{S}_{t}$ and the observable state of the $i$-th human at time $t$  $\boldsymbol{O}_{t}^{i}$ become:
\begin{equation}
\label{states}
  \begin{split}
  \boldsymbol{S}_{t} &= [d_{g}, v_{pref}, v_{x}, v_{y}, r], \\
  \boldsymbol{O}_{t}^{i} &= [d^{i},p_{x}^{i}, p_{y}^{i}, v_{x}^{i}, v_{y}^{i}, r^{i}, r^{i}+r].
  \end{split}
\end{equation}

The joint state of all $(n+1)$ agents at time $t$ can be obtained by concatenating the state of the robot with the observable states of humans as $\boldsymbol{J}_{t} = [\boldsymbol{S}_{t}, \boldsymbol{O}_{t}^{1}, \boldsymbol{O}_{t}^{2},\ldots, \boldsymbol{O}_{t}^{n}]$. Assume that the robot can change its velocity immediately according to the action command at time $t$ $\boldsymbol{a}_{t}$, which is determined by the navigation policy: $\boldsymbol{v}_{t}=\boldsymbol{a}_{t} = \boldsymbol{\pi}(\boldsymbol{J}_{t})$. The corresponding reward at time $t$ is denoted as $R(\boldsymbol{J}_{t},\boldsymbol{a}_{t})$. The definition of reward function in \cite{cadrl}\cite{chen2017socially}\cite{sarl} is shown in (\ref{reward}), where $d_{min}$ is the shortest separation distance between the robot and humans within the decision interval $\Delta t$, and $d_{c}$ represents the minimum comfortable distance that humans can tolerate.
\begin{equation}
\label{reward}
R(\boldsymbol{J}_{t},\boldsymbol{a}_{t}) =
  \begin{cases}
  -0.25, & \text{if } d_{min} < 0; \\
  0.5 * (d_{min}-d_{c}), & \text{if } 0 < d_{min} < d_{c};\\
  1, & \text{if } d_{g} = 0; \\
  0, & \text{otherwise}. \\
  \end{cases}
\end{equation}
Equation (\ref{reward}) indicates that actions that lead the robot to the goal will be awarded while actions that may cause collision or discomfort to pedestrians will be penalized. If the robot satisfies any of the following conditions: a) reaches the destination (i.e., the position is close enough to the goal), b) collided with a human, and c) the navigation time exceeds the time limit, the episode will be terminated.

With an optimal navigation policy $\boldsymbol{\pi}^\star(\boldsymbol{J}_{t})$, the optimal value of the joint state $\boldsymbol{J}_t$ at time $t$ can be formulated as:
\begin{equation}
\label{value}
  V^\star(\boldsymbol{J}_{t})= \sum_{i=0}^K \gamma^{i \cdot \Delta t \cdot v_{pref}} \cdot R(\boldsymbol{J}_{t},\boldsymbol{a}_t^\star),
\end{equation}
where $K$ is the total number of decision steps from the state at time $t$ to the final state, $\Delta t$ is the decision interval between two actions, and $\gamma\in(0,1)$ is a discount factor in which the preferred speed $v_{pref}$ is introduced as a normalization parameter for numerical reasons \cite{cadrl}.
The optimal policy is formulated by maximizing the cumulative reward as the following:

\begin{small}
\begin{equation}
\label{policy1}
\begin{split}
  \boldsymbol{\pi}^\star(\boldsymbol{J}_{t}) =&\mathop{\arg\max}_{\boldsymbol{a}_t\in\boldsymbol{A}} R(\boldsymbol{J}_{t},\boldsymbol{a}_t)+ \gamma^{\Delta t \cdot v_{pref}} \cdot \\
  &\int_{\boldsymbol{J}_{t+\Delta t}} P(\boldsymbol{J}_{t+\Delta t} \mid \boldsymbol{J}_{t},\,\boldsymbol{a}_t) \cdot V^\star(\boldsymbol{J}_{t+\Delta t}) \, \mathrm{d} \boldsymbol{J}_{t+\Delta t},
\end{split}
\end{equation}
\end{small}
where $\boldsymbol{A}$ is the action space (i.e., the set of velocities that can be achieved), $P(\boldsymbol{J}_{t+\Delta t}\mid\boldsymbol{J}_{t},\,\boldsymbol{a}_t)$ is the transition probability from $\boldsymbol{J}_{t}$ to $\boldsymbol{J}_{t+\Delta t}$ when action $\boldsymbol{a}_t$ is carried out. This probability describes the uncertainty of the next joint state due to the unknown crowd behavior. 

\section{METHODS}

\subsection{Background}

\subsubsection{Value Function Approximation (VFA)}

In our study, we adopt the Value Function Approximation (VFA) algorithm as the basic RL algorithm to solve our task. In detail, we use a deep neural network with the architecture proposed in \cite{sarl} to approximate the optimal value function $\hat{V}(\boldsymbol{J}_{t})$. In the policy evaluation, the dynamics of the environment can be considered accessible, i.e., it is assumed that the next joint state of the robot and humans $\boldsymbol{J}_{t+1}$ after the robot takes an action at the current state $\boldsymbol{J}_{t}$ can be approximated using some predictive models \cite{sarlstar}: $\boldsymbol{J}_{t+1} \leftarrow propogate(\boldsymbol{J}_{t}, a_t)$. Then the optimal policy can be retrieved from the optimal value function $V^\star(\boldsymbol{J}_{t+\Delta t})$. By this simplification, the calculation of the optimal policy in (\ref{policy1}) can be modified as (\ref{policy2}). 

\begin{small}
\begin{equation}
\label{policy2}
  \boldsymbol{\pi}^\star(\boldsymbol{J}_{t}) = \mathop{\arg\max}_{\boldsymbol{a}_t\in\boldsymbol{A}} R(\boldsymbol{J}_{t},\boldsymbol{a}_t)+ \gamma^{\Delta t \cdot v_{pref}} \cdot \hat{V}(\boldsymbol{J}_{t+\Delta t}).
\end{equation}
\end{small}

In this work, the pedestrians are assumed to move with constant velocities during the time interval $[t, t+\Delta t]$, as the duration $\Delta t$ is very small. Therefore, by using a constant velocity model (CVM), the next joint state of robot and humans can be predicted: $\boldsymbol{J}_{t+\Delta t}\leftarrow \text{CVM}(\boldsymbol{J}_{t},\Delta t,\boldsymbol{a}_t)$. 

The value network is trained by the temporal-difference (TD) method with experience replay and fixed target network techniques.

\subsubsection{Hindsight Experience Replay (HER)}

In our problem formulation, the defined reward function (\ref{reward}) is sparse, which is difficult for the agent to obtain non-trivial rewards with random explorations. Dealing with sparse rewards is always more challenging in RL. The hindsight experience replay technique (HER) proposed in \cite{andrychowicz2017hindsight} is a promising method to solve the sparse reward problems. The key insight of HER is that the agent can learn useful information from the failed rollouts by viewing the final state as its additional goal, as if the agent intended on reaching this state from the very beginning. The idea is realized by relabelling the failed experiences as successful ones. The details of our RL+HER algorithm in our task is outlined in Algorithm 1. 

For each episode, if there is a collision or the agent achieves the goal, we will directly store the trajectory in the experience replay buffer; if the final state of the robot is ``Timeout" without causing discomfort to humans, we will relabel the final state as the goal and the last reward as a successful reward and then store the modified trajectory in the replay buffer. HER is a simple method without complicated reward engineering and can help improve the sample efficiency in RL.

\begin{algorithm}[tb]  \small
  \caption{VFA Learning with HER}  
  \begin{algorithmic}[1] 
    \State Initialize value network $V$ and target value network $\hat V$
    \State Initialize experience replay memory $E$
    \For{episode $i\in [1,M]$}  
      \State Sample an initial state $\boldsymbol{J}_{0}$ with the original goal $\boldsymbol{g}$
       \For{$t=0$ , $T-1$ }  
      	\State $\boldsymbol{a}_t \leftarrow \boldsymbol{\pi}^\star(\boldsymbol{J}_{t}) = \mathop{\arg\max}_{\boldsymbol{a}_t\in\boldsymbol{A}} R(\boldsymbol{J}_{t},\boldsymbol{a}_t)+ \gamma^{\Delta t \cdot v_{pref}} \cdot \hat{V}(\boldsymbol{J}_{t+1})$
      	\State Execute the action $\boldsymbol{a}_t$ and observe a new state $\boldsymbol{J}_{t+1}$
      \EndFor
      \State Record information $\textit{info}$ of the last state $\boldsymbol{J}_{T}$ 
      \If {$\textit{info} = \textit{ReachGoal}$ or $\textit{Collision}$} 
        \For{$t=0$ , $T-1$ } 
          \State Store the transition $(\boldsymbol{J}_{t}, \boldsymbol{a}_t, \boldsymbol{r}_t, \boldsymbol{J}_{t+1})$ in $E$
        \EndFor  
      \ElsIf {$\textit{info} = \textit{Timeout}$}
	  \State Relabel the final agent position as the addition goal:
	  \State $\boldsymbol{g'} \leftarrow \boldsymbol{p}_T$ 
	\For{$t=0$ , $T-1$ } 
	  \State Obtain $\boldsymbol{J'}_{t}$ and $\boldsymbol{J'}_{t+1}$ with $\boldsymbol{g'}$
	  \If {$\boldsymbol{p}_t = \boldsymbol{g'}$}
	    \State $\boldsymbol{r'}_t = 1$
	  \Else
	    \State $\boldsymbol{r'}_t = \boldsymbol{r}_t$
	  \EndIf 
	   \State Store the transition $(\boldsymbol{J'}_{t}, \boldsymbol{a}_t, \boldsymbol{r'}_t, \boldsymbol{J'}_{t+1})$ in $E$
	\EndFor 
	\EndIf
	\For{$t=0$ , $N$ } 
	  \State Sample a minibatch $B$ from the replay buffer $E$
	  \State Set target $y_i = \boldsymbol{r}_i + \gamma^{\Delta t \cdot v_{pref}} \cdot \hat{V}(\boldsymbol{J}_{i+1})$
	  \State Update value network $V$ by gradient descent
	\EndFor 
	\If {$\text{episode } \% \text{ target update interval} = 0$}
	  \State Update target network $V \leftarrow \hat{V}$
	\EndIf
    \EndFor  
    \label{code:recentEnd}  
  \end{algorithmic}  
\end{algorithm}

\subsection{Reward shaping (RS)}

To deal with the sparse reward task of human-aware navigation without employing expert demonstrations, the most intuitive method is to shape the reward function. Using reward shaping (RS) can be beneficial with carefully engineering. Since the task of robot navigation is to reach a given goal, we incorporate an distance-to-goal reward to the original sparse reward (\ref{reward}). Since simply using distance-to-goal to shape the reward often fails as it renders learning vulnerable to local optima, it needs to be well designed for a successful application. We shape the reward only based on the distance-to-goal function without any additional domain knowledge, as shown in (\ref{reward1}):
\begin{equation}
\label{reward1}
R(\boldsymbol{J}_{t},\boldsymbol{a}_{t}) =
  \begin{cases}
  -1, & \text{if } d_{min} < 0; \\
  0.5 * (d_{min}-d_{c}), & \text{if } 0 < d_{min} < d_{c};\\
  2, & \text{if } d_{g} = 0; \\
  -\alpha * dist, & \text{otherwise}. \\
  \end{cases}
\end{equation}
where $dist =\|\boldsymbol{p}-\boldsymbol{g}\|$ is the distance from the agent to the fixed goal and $\alpha$ is a hyper-parameter. In our task, we set $\alpha = 0.002$ considering the relative magnitudes between the distance value and the other reward values and having several experimental trials.

\subsection{Curriculum Learning (CL) with HER}
Although the RS method can provide a dense reward, it usually needs very careful engineering to achieve good performance. Therefore, we propose a another method to deal with the sparse reward problem. As illustrated in Algorithm 1, we first implement the HER technique to relabel the final state of the ``Timeout" trajectories as goals and store them in the experience replay. Therefore, the agent can also learn from the failure.

In practice, we find that when there are many humans in the simulated environment and the punishment for collision is very high (-1), the agent can hardly get successful trajectories during training, even equipped with the HER technique. Therefore, we introduce the curriculum learning (CL) in our method, to make the agent start out with easy tasks and then gradually increase the task difficulty. In our implementation, we first train the agent in a simple environment that contains only one human to make it learn to navigate without collision, and then transfer it to a more complex environment with $5$ humans.

\begin{table}[bt] \renewcommand\arraystretch{1.1}
\setlength{\abovecaptionskip}{-0.1cm}
\centering
\caption{Navigation Performance of RL Agents Trained With and Without Using Demonstration Data}
\begin{tabular}{p{0.04\textwidth}<{\raggedright}p{0.08\textwidth}<{\raggedright}p{0.04\textwidth}<{\raggedright}p{0.05\textwidth}<{\raggedright}p{0.04\textwidth}<{\raggedright}p{0.03\textwidth}<{\raggedright}p{0.04\textwidth}<{\raggedright}} 
\toprule
Demo &Method&Success &Collision &Danger & Time& Reward\\
  \midrule
\multirow{3}{*}{Use} & RL	&1.00&0.00&0.47&10.58&0.3342  \\
   &   RL+IL&1.00&0.00	&0.23	&10.72&	0.3320\\
	& RL+HER	&\textbf{1.00}	&\textbf{0.00}&	\textbf{0.17}&	\textbf{10.57}&	\textbf{0.3370} \\
\hline
\multirow{3}{*}{Not use}&RL	&0.00	&\textbf{0.00}	&\textbf{0.00}	&--	&0.0000 \\
	&RL+RS	&0.78	&0.22	&1.23	&\textbf{11.48}	&0.1420 \\
	&RL+HER+CL	&\textbf{0.97}	&0.03	&0.27	&12.32	&\textbf{0.2642}\\
  \bottomrule
  \end{tabular}
  \label{performance}
  \vspace{-0.3cm}
\end{table}

\section{EXPERIMENTS}

\subsection{Simulation Environment}

The simulation environment for human-aware navigation is built in Python. There are $5$ humans in the simulated crowd, whose navigation policy is determined by the optimal reciprocal collision avoidance (ORCA) method \cite{van2011reciprocal}. We use circle crossing scenarios, where the positions of the humans and the robot are randomly initialized on a circle of radius $4m$ and their goal positions are on the opposite side of the circle. We assume holonomic kinematics for the robot, i.e., it can move in any direction. The action space consists of $9$ discrete actions in total: the stop action and the speed of $1m/s$ with $8$ heading directions evenly spaced between $[0, 2\pi)$.

\begin{figure}[tb]
\setlength{\abovecaptionskip}{-0.1cm}
      \centering
      \includegraphics[scale=1.0,angle=0,width=0.48\textwidth]{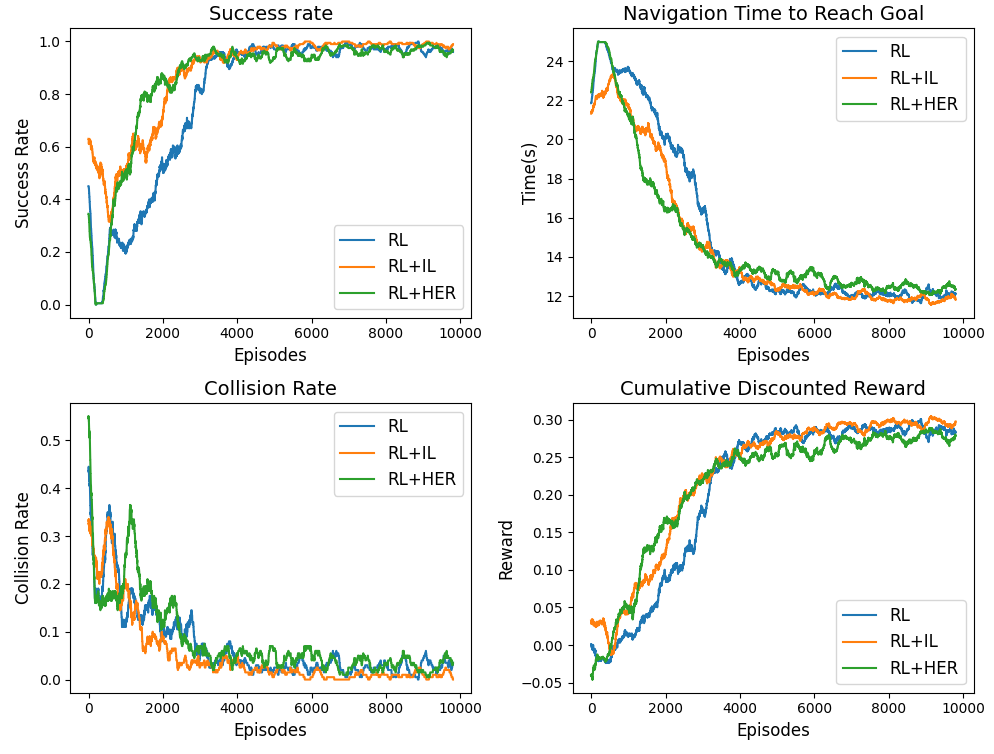}
      \caption{Learning curves of the RL, RL+IL, RL+HER methods when expert demonstration data is used in training.}
      \label{demo}
      \vspace{-0.2cm}
\end{figure}

\subsection{Evaluation of Agents Trained with Expert Demonstrations}
First of all, we evaluate the agents trained with expert demonstrations. Specifically, we first implement the baseline \cite{sarl} that uses demonstration data to initialize the experience replay memory and combines IL+RL to train the agent. Then we implement RL with initialized memory without using IL for pre-training. Finally, we implement RL+HER with initialized memory but does not use IL for pre-training. By designing these experiments, we want to study the separate contributions of IL and HER to the learned navigation policy when the demonstration data is used. 

The demonstration data is generated by navigating with ORCA policy for $3000$ episodes. For imitation learning, we pre-train the value network for $50$ epochs on the demonstration data with a learning rate of $0.01$. In the reinforcement learning phase, we apply a learning rate of $0.001$ and an $\varepsilon$-greedy policy for action selection. The exploration rate of the $\varepsilon$-greedy policy decays linearly from $0.5$ to $0.1$ in the first $5000$ episodes and stays at $0.1$ for the remaining episodes. The value network is trained for $10000$ RL episodes. The training set and validation set use different seeds for initialization of the positions of the agents and simulated humans. Each agent is tested in $500$ episodes after training.

The navigation performance of the RL, RL+IL and RL+HER agents trained with demonstration data are shown in Table \ref{performance} (first three lines) and Fig. \ref{demo}. It can be seen that all three methods can obtain a success rate of 1 with zero collision. However, the RL+HER agent achieves the lowest frequency of being in danger (the number of times the robot is too close to human divided by the total number of episodes), the shortest time to reach the goal, and the highest discounted cumulative reward among the agents. Also, as shown in Fig. \ref{demo}, the use of HER can help improve the sample efficiency and accelerate the learning process. The results demonstrate that our method can better learn from failure to achieve a better performance.

\subsection{Evaluation of Agents Trained without Using Expert Demonstrations}

\begin{figure}[tb]
\setlength{\abovecaptionskip}{-0.1cm}
      \centering
      \includegraphics[scale=1.5,angle=0,width=0.49\textwidth]{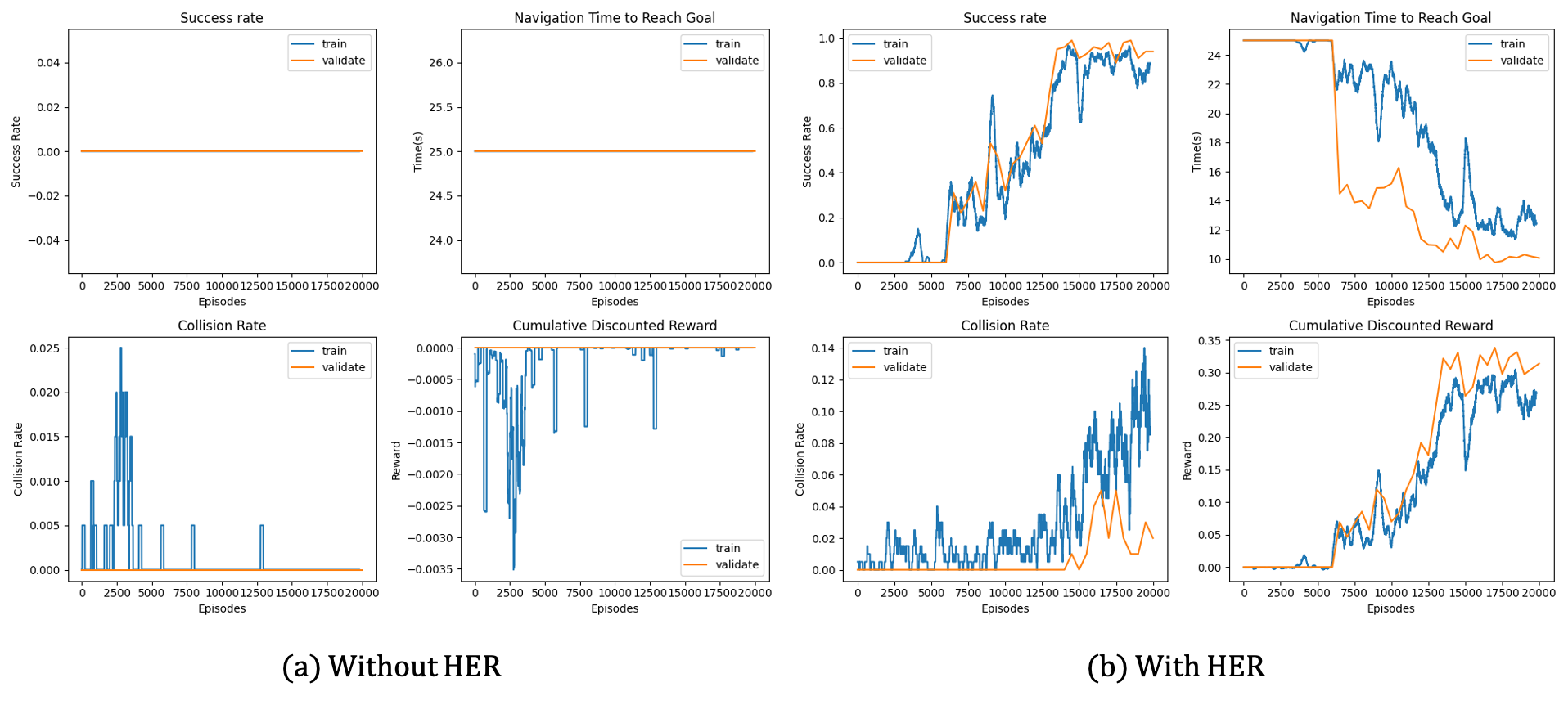}
      \caption{Learning curves of the RL agents (a) without HER  and (B) with HER, in an environment with 1 human. Expert demonstration data is not used in training.}
      \label{h1}
\end{figure}

Then, we conduct the navigation experiments without employing demonstration data. Instead, we initialize the experience replay memory with $3000$ episodes generated with random actions. Other implementation details are the same as those in Section~IV-B.

First, we compare the RL agents with and without using HER in the navigation task in an environment with only 1 human. The learning curves are compared in Fig. \ref{h1}. As shown in Fig. \ref{h1}, the RL agent without HER can learn to avoid collision with human, but never gets successful experiences. This is because the reward is too sparse and there is no expert demonstration as a guidance. Hence, it is difficult to learn a successful navigation policy even in a simple environment with only 1 human. However, with the introduction of HER, the experience replay memory becomes more informative and enables the successful learning process, as shown in Fig. \ref{h1}(b). During training, the success rate gradually increases and converges to nearly $100\%$ with a low collision rate. The experimental result proves the effectiveness of HER algorithm in dealing with sparse reward problems, without any additional domain expertise.

\begin{figure}[tb]
\setlength{\abovecaptionskip}{-0.1cm}
      \centering
      \includegraphics[scale=1.0,angle=0,width=0.49\textwidth]{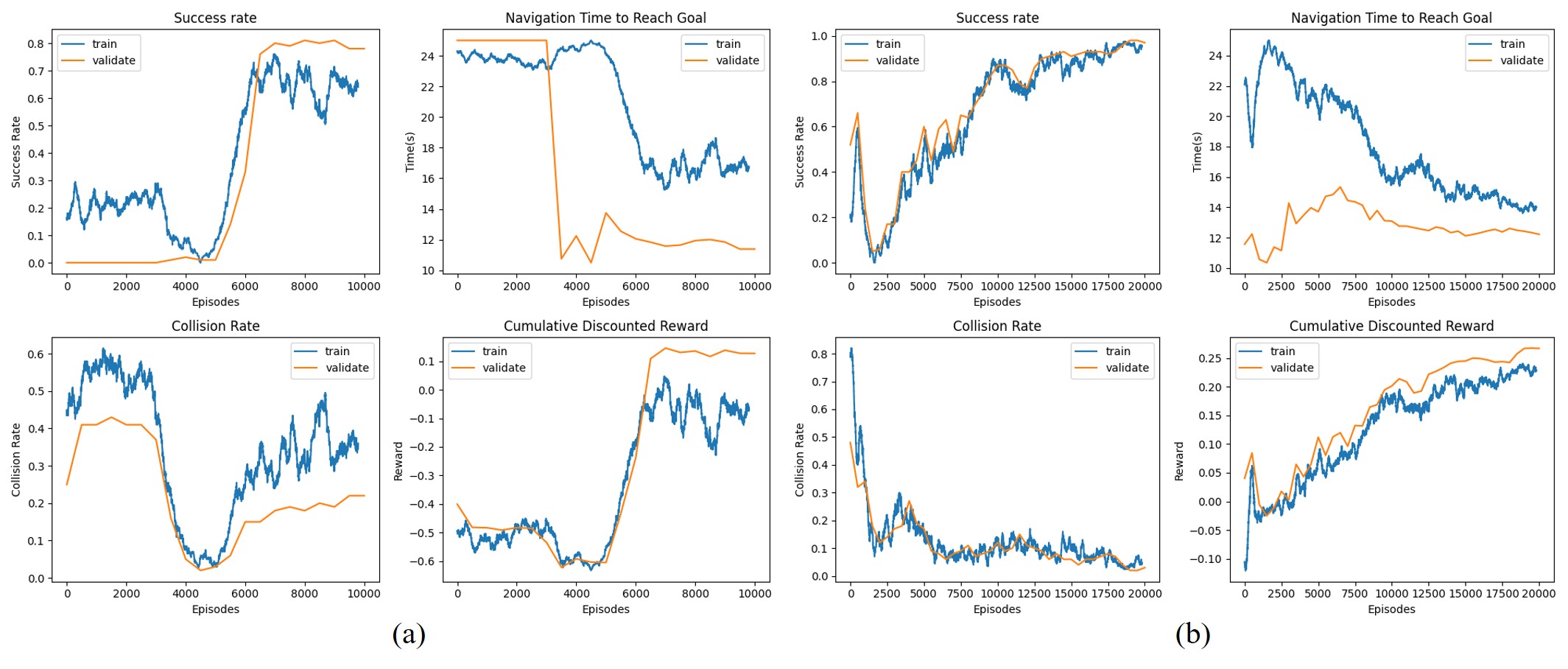}
      \caption{Learning curves of the (a) RL+RS and (b) RL+HER+CL agents, in an environment with 5 humans. Expert demonstration data is not used in training.}
      \label{cl}
\vspace{-0.2cm}
\end{figure}

\begin{figure*}[tb]
\setlength{\abovecaptionskip}{-0.1cm}
      \centering
      \includegraphics[scale=1.0,angle=0,width=0.97\textwidth]{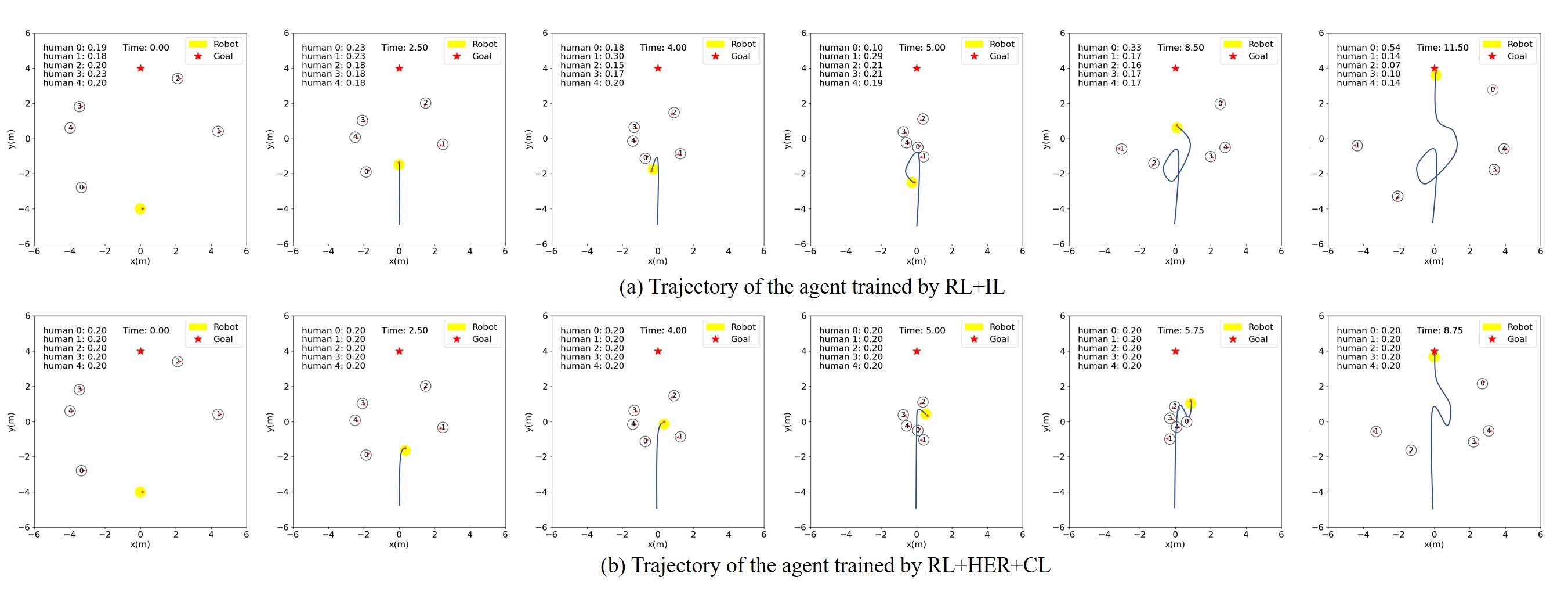}
      \caption{Trajectories of the agents trained by (a) the baseline method RL+IL, and by (b) our proposed RL+HER+CL method in an environment with 5 humans. Our method outperforms the baseline in both the travel time and the trajectory length.}
      \label{traj}
\vspace{-0.2cm}
\end{figure*}

We then evaluate the RL methods (i.e., RL, RL+RS and RL+HER+CL) in a more difficult environment that contains the robot and a crowd of $5$ humans. For the RL+HER+CL method, we initialize the value network using the learned weights in the navigation with $1$ human, and train the agent in an environment with $5$ humans.
The results are presented in Table \ref{performance} (last three lines). As expected, the RL agent directly trained in the difficult sparse-reward environment can never achieve successful results and only learns to avoid humans. The RL+RS agent can successfully navigate to the goal in most of the cases ($78\%$) with a shortest navigation time, but also has a high collision rate ($22\%$). While the RL+HER+CL method can learn to navigate towards the goal with a high successful rate ($97\%$) and a low collision rate ($3\%$) with the highest total reward among all methods, and the frequency of being in danger is also the lowest, which demonstrates the superiority of our method.

The learning curves of the RL+RS and RL+HER+CL agents are further compared in Fig. \ref{cl}. It can be seen in Fig. \ref{cl}(a) that for both training and validation sets, the cumulative discounted reward gradually increases during the training process, and the success rate and navigation time of the agent improve a lot. This means the method using shaped reward can successfully learn the navigation actions to reach the goal in the crowd-navigation scenarios. However, the collision rate first decreases in the first $5000$ training episodes, but increases in the remaining $5000$ episodes. This may be due to our self-designed reward function, which requires more careful engineering. The results reveal that although the shaped reward is intuitive and simple, it needs a large amount of engineering effort and might not be useful for the complex crowd navigation task. 
However, as shown in Fig. \ref{cl}(b), the success rate of the RL+HER+CL agent gradually converges to about $100\%$ with almost zero collision rate. The results show that our method can successfully learn a human-aware navigation policy in the dense crowds without employing expert demonstrations and achieves a comparative performance with the state-of-the-art RL+IL methods that use expert demonstrations. This method can be easily applied in real world navigation problems, as it does not require additional effort for the collection of human demonstration data.

Furthermore, we compare the performance of our RL+HER+CL method with the baseline method (i.e., RL+IL) in a test case. As shown in Fig. \ref{traj} (a), although the demonstration data is employed in the training of the agent in baseline, it may not always make optimal and efficient decisions to navigate through a crowd towards its goal. It makes detours between the 4th to 5th seconds, and travels a longer distance to achieve the goal. In contrast, as shown in Fig. \ref{traj} (b), without using expert demonstration data for training, our RL+HER+CL agent can learn to efficiently arrive at the goal with much lower time consumption and a shorter trajectory length. The results demonstrate the effectiveness of our proposed approach. 

\subsection{Demonstration Video}
The demonstration video can be found at \url{https://youtu.be/QaYlXBJPnyI} for a better visualization of the results.

\section{CONCLUSIONS}
In this paper, we study the human-aware robot navigation problem via RL, and propose a framework to incorporate the HER and CL techniques, to learn both efficient and safe navigation in a dense crowd without using additional expert demonstration data. The goal-oriented task is formulated as a sparse reward problem. Without requiring demonstration data for model initialization, we develop two different methods, i.e., RL+RS and RL+HER+CL, to address the sparse reward navigation problem. The experiment results show that our proposed RL+HER+CL method can successfully learn the human-aware navigation policy and can achieve even better performance in terms of time consumption and trajectory length compared with the state-of-the-art methods that use additional demonstration data in training.


\addtolength{\textheight}{-1cm}   % This command serves to balance the column lengths


%%%%%%%%%%%%%%%%%%%%%%%%%%%%%%%%%%%%%%%%%%%%%%%%%%%%%%%%%%%%%%%%%%%%%%%%%%%%%%%%
% USE THIS reference format
\bibliographystyle{ieeetr}   
\bibliography{wcica_latex_sample}
\end{document}